\documentclass[11pt]{article}
\usepackage{acl}
\usepackage{times}
\usepackage{latexsym}
\usepackage{microtype}
\usepackage{inconsolata}
\usepackage{graphicx}
\usepackage{booktabs}
\usepackage{amsmath,amssymb}
\usepackage{array}
\usepackage{tabularx}
\usepackage[T1]{fontenc}
\usepackage{placeins}
\usepackage{needspace}
\title{Auditing Current-Vector Selectivity in Cross-Model Residual Transfer}
\author{Jiawei Xie\textsuperscript{1,2,3*} \quad Gongce Cao\textsuperscript{1,2,3*} \quad Enmeng Lu\textsuperscript{2} \quad Yu Chen\textsuperscript{1,2,3} \quad Yuxuan Peng\textsuperscript{1,2,3} \\
  \textbf{Mingyang Lyu\textsuperscript{2,1,4} \quad Jinyu Fan\textsuperscript{2} \quad Shuai Yang\textsuperscript{2} \quad Zhengqi Wang\textsuperscript{2} \quad Yi Zeng\textsuperscript{3,2\dag}} \\[2pt]
  \normalsize\mdseries \textsuperscript{1}University of Chinese Academy of Sciences \quad \textsuperscript{2}Beijing Institute of AI Safety and Governance \\
  \normalsize\mdseries \textsuperscript{3}Gaoling School of AI, Renmin University of China \quad \textsuperscript{4}Zhongguancun Academy \\[1pt]
  \small\mdseries \textsuperscript{*}Equal contribution. \quad \textsuperscript{\dag}Corresponding author: \texttt{yi.zeng@ruc.edu.cn}}
\begin{document}
\maketitle
\begin{abstract}

Cross-model residual interventions can change a receiver's output without showing that the transferred vector is specific to the current instance. We develop matched other-instance substitution, an oracle-oriented receiver-side diagnostic that compares the current vector with an equal-norm donor while keeping answer orientation and the receiver interface fixed. The current-minus-donor contrast measures current-vector selectivity relative to the assigned donor. We test five open-weight checkpoints in two overlapping three-model systems on Moral Stories and \mbox{XStoryCloze}. On Moral Stories, \mbox{HET-3} current vectors increase gold margin by 1.913 (simultaneous 95\% CI [1.370, 2.455]) and accuracy by 8.63 percentage points relative to matcher-nearest same-answer donors. In a prespecified \mbox{XStoryCloze} study, Qwen-aligned current vectors yield a smaller +0.226 margin effect (simultaneous 95\% CI [0.082, 0.370]) over position-matched, content-similar donors, without an accuracy gain. The contrast separates decision-level current-vector effects from graded margin shifts at a fixed residual interface.

\end{abstract}

\section{Introduction}
\label{sec:introduction}

\begin{figure*}[t]
\centering
\includegraphics[width=\textwidth]{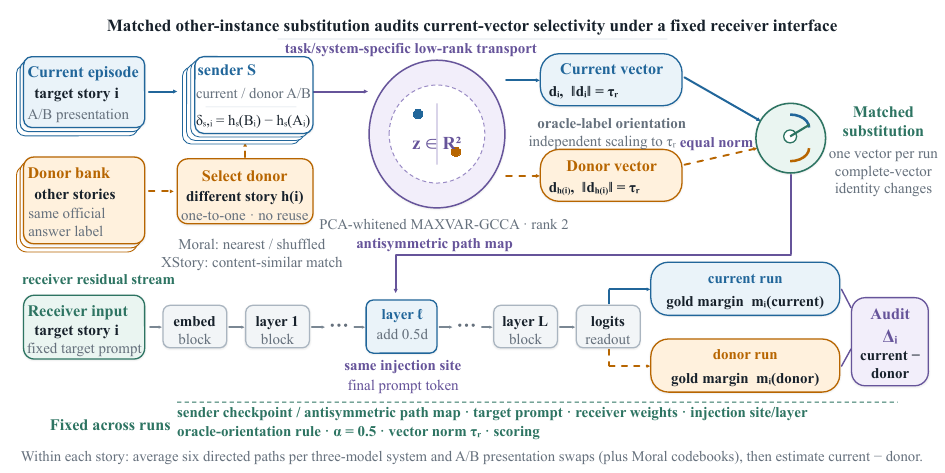}
\caption{Matched other-instance substitution at a fixed receiver interface. A task-specific PCA-whitened rank-2 map transports current vector \(d_i\) and answer-label-matched donor vector \(d_{h(i)}\) into the same receiver. Both vectors are independently oracle-oriented and rescaled to target norm \(\tau_r\). The receiver input, prompt, injection site, layer, coefficient, patch norm, and scoring rule remain fixed; only complete-vector identity changes. The contrast \(\Delta_i=m_i(d_i)-m_i(d_{h(i)})\) measures current-vector value relative to the assigned donor.}
\label{fig:protocol}
\end{figure*}

Continuous interfaces between language models can transport hidden states, residual directions, and key--value caches. A transferred state may reflect the current input or supply a generic direction toward an answer already recoverable from the prompt. Zero and random-vector controls measure intervention efficacy; determining current-instance value requires comparison with another plausible complete vector.

Recent causal audits compare current-example, other-example, no-message, and self-generated messages; same-answer controls have also been used to test activation transduction \citep{zhang2026causalaudit,quirke2026ablation}. Interlat includes CrossTask and structured random controls for learned latent channels \citep{du2026interlat}. In cross-model residual transfer, vector magnitude can differ across checkpoints, and candidate order can reverse a paired direction. Our audit independently norm-matches complete vectors, assigns one answer-label-matched donor to each target, and enforces antisymmetric transport under candidate swaps. Table~\ref{tab:closest-audits} compares this estimand with the closest controls.

We ask whether, under one receiver interface and intervention norm, the current transported vector produces a larger gold margin than a complete vector from another answer-label-matched instance. The donor preserves label orientation and injection mechanics while changing vector identity. We call this current-minus-donor contrast \emph{current-vector selectivity}: the receiver-side compatibility of the current vector relative to the assigned donor. Figure~\ref{fig:protocol} shows the substitution and the quantities held constant.

We evaluate five publicly released checkpoints in two overlapping three-model alignment systems. Each system fits a task-specific rank-2 alignment and evaluates all six directed sender--receiver paths. Moral Stories yields a large \mbox{HET-3} current-minus-donor effect accompanied by decision changes. A separate 96-story \mbox{XStoryCloze} study yields a smaller Qwen-aligned gold margin effect without an accuracy gain. The contrast distinguishes current-vector value from generic intervention efficacy in both task environments.

We make two contributions:

\begin{enumerate}

\item \textbf{A matched-interface estimand for cross-model residual transfer.} Complete current and donor vectors share label orientation, receiver prompt, intervention site, scoring rule, and patch norm. One-to-one assignment and antisymmetric transport preserve the comparison under candidate-order changes.

\item \textbf{A two-task characterization of graded and decision-level effects.} Moral Stories links the current-minus-donor margin to decision changes; \mbox{XStoryCloze} yields a smaller graded effect while accuracy remains unchanged.

\end{enumerate}

\section{Related Work}
\label{sec:related-work}

\subsection{Latent communication and cross-model transfer}

Activation Communication combines one agent's activation with another agent's hidden state, while Interlat learns continuous channels between agents, including heterogeneous-model settings \citep{ramesh2025communicating,du2026interlat}. Other work transports residuals, concepts, or KV caches across architectures and studies linear transferability, compression, and real-time interaction \citep{huang2025crossmodel,chen2026dense,li2026latentbridge,wenzel2026latent}. These studies show that continuous states can alter another model's output. We test whether the vector computed for the current instance has value beyond an answer-label-matched complete-vector substitute.

Concurrent causal audits compare current-example, other-example, no-message, and self-generated latent messages across reasoning tasks \citep{zhang2026causalaudit}. Same-answer controls also appear in stress tests of activation transduction \citep{quirke2026ablation}. Our cross-model residual audit independently norm-matches complete vectors and assigns one label-matched donor per target. It combines these controls with antisymmetry under presentation swaps and story-level simultaneous inference. Table~\ref{tab:closest-audits} compares the resulting estimands.

\begin{table}[htbp]
\centering
\footnotesize
\setlength{\tabcolsep}{2.5pt}
\begin{tabular}{@{}>{\raggedright\arraybackslash}p{0.18\linewidth}>{\raggedright\arraybackslash}p{0.77\linewidth}@{}}
\toprule
Study & Identification design and estimand \\
\midrule
\citet{du2026interlat} (Interlat) & Continuous inter-agent channel with CrossTask and structured random controls; compares matched-task messages with mismatched-task or synthetic messages. \\
\citet{zhang2026causalaudit} & Current, other-example, no-message, and self-generated substitutions; separates message presence, example identity, and sender-specific content. \\
\citet{quirke2026ablation} & Activation transduction with a same-answer control; tests whether mechanistic-role claims survive ablation and cross-model transduction. \\
This work & Cross-model residual injection with independently equal-norm complete vectors, one-to-one label-matched donors, a fixed receiver interface, and exact order antisymmetry; estimates current-vector value relative to the assigned donor. \\
\bottomrule
\end{tabular}
\caption{Comparison with the closest latent-channel and substitution audits. Our design jointly fixes answer orientation, complete-vector norm, receiver interface, and presentation-order behavior while varying vector identity.}
\label{tab:closest-audits}
\end{table}

\subsection{Activation steering and causal intervention}

Contrastive activation addition, conditional steering, and activation transport show that low-dimensional directions can systematically move model behavior \citep{turner2023activationengineering,rimsky2024steering,lee2025cast,rodriguez2025activationtransport}. Activation patching further emphasizes that causal conclusions depend on the intervention site, scale, corruption, and metric \citep{zhang2024activationpatching,geiger2025causalabstraction}. A transported vector and a steering vector can be identical at the receiver; their distinction lies in what changes across inputs. Answer-label-matched substitution preserves shared orientation and injection mechanics while varying vector identity.

\subsection{Representation alignment and readout dependence}

CCA-family measures, centered kernel alignment, and model stitching quantify or operationalize representational correspondence \citep{kettenring1971canonical,kornblith2019similarity,bansal2021revisiting}. Representation similarity can coexist with different functional use, motivating intervention-based evaluation \citep{huh2024platonic,smith2025functional,athanasiadis2026grounding}. Because the shared coordinate is learned by construction, geometry defines the interface rather than the outcome. We evaluate that interface with a matched functional contrast.

Language-model measurements also vary with verbalizers and answer surface forms \citep{jiang2020know,holtzman2021surfaceform,li2024verbalizer}. We therefore keep the readout fixed inside each contrast, counterbalance candidate order, and distinguish continuous gold margin from accuracy and free generation.

\section{Matched Other-Instance Substitution}
\label{sec:method}

The audit changes one element of the receiver-side computation: the identity of the complete intervention vector. Figure~\ref{fig:protocol} traces the current and donor arms through the same transport, normalization, injection site, and readout.

\subsection{Task-specific low-rank transport}
\label{sec:transport}

For story \(i\), sender \(s\), and two displayed candidates \(A\) and \(B\), we extract paired hidden-state differences at preselected early and late views:

\[
\delta_{s,i}=h_s(x_i^B)-h_s(x_i^A).
\]

For each three-model system, we fit a six-view maximum variance generalized canonical correlation analysis (MAXVAR-GCCA) state. The views are each model's early and late differences after principal component analysis (PCA) whitening. Moral Stories uses 512 formation pairs; \mbox{XStoryCloze} uses 256 stories from the official training split. The shared space has rank 2. For sender--receiver path \(s\rightarrow r\), we antisymmetrize the uncentered map:

\[
A_{s\rightarrow r}(\delta)=\tfrac12\left[F_{s\rightarrow r}(\delta)-F_{s\rightarrow r}(-\delta)\right].
\]

The diagnostic uses the official answer label to orient both current and donor directions, then independently rescales each direction to the receiver's formation-only target norm \(\tau_r\):

\[
d_{s\rightarrow r,i}=\sigma_i\tau_r\frac{A_{s\rightarrow r}(\delta_{s,i})}{\lVert A_{s\rightarrow r}(\delta_{s,i})\rVert_2}.
\]

Here \(\sigma_i\in\{-1,+1\}\) is the official-label orientation; each donor is oriented by its own label. The target norm \(\tau_r\) is the median L2 norm of receiver late-view differences on formation data (Appendix~\ref{app:target-norms}). We inject \(\alpha d_i\), with \(\alpha=0.5\), at the preselected receiver late layer and final prompt token. Candidate-order swaps negate the displayed difference and its orientation, preserving the effective patch. The executed uncentered map is linear, so \(A=F\); we verify this identity under swaps.

\subsection{Complete-vector substitution}
\label{sec:substitution}

Let \(h(i)\) be a different story with the same official answer label. The current arm applies \(d_i\); the donor arm applies \(d_{h(i)}\). Each nonzero complete direction is independently rescaled to \(\tau_r\), so the comparison changes vector identity rather than intervention magnitude. The primary outcome is the task-native gold margin:

\[
m_i(v)=\log p_r(y_i^+\mid c_i,v)-\log p_r(y_i^-\mid c_i,v).
\]

Here \(c_i\) is the target prompt, and \(y_i^+\) and \(y_i^-\) are the correct and incorrect answer labels. The principal estimand is \(\theta=\mathbb{E}_i[m_i(d_i)-m_i(d_{h(i)})]\).

Moral Stories also includes a shuffled same-answer donor, exact zero, and a cross-fitted oracle generic direction. The target-conditioned decomposition is exact by construction: for target-specific generic axis \(g_i\), \(e_{j\mid i}=d_j-g_i\), hence \(g_i+e_{j\mid i}=d_j\). The complete current and donor arms therefore remain equal-norm vectors; the generic-only arm measures a separate steering effect.

\subsection{Donor construction}
\label{sec:donors}

Moral Stories donors are selected within answer class by a predefined TF-IDF matcher over structured story fields, with a one-to-one Hungarian assignment and no fixed points. We refer to this comparator as the \emph{matcher-nearest same-answer donor}. Its mean matcher score is 0.166, compared with 0.113 for the shuffled same-answer donor.

\mbox{XStoryCloze} uses only receiver-visible text---the four-sentence context and both candidate endings---embedded with a versioned \texttt{all-mpnet-base-v2} checkpoint. The 96 targets and 96 donors come from disjoint story sets. Donors are assigned globally without reuse or outcome access. Matching uses the source record's correct candidate position (1 or 2), not a semantic response category. Canonical and swapped presentations counterbalance that position at scoring time. Assigned cosine similarity has mean 0.658, median 0.669, and range 0.509--0.844. An outcome-blind single-rater audit of 16 pairs used an event/causal-overlap rubric and rated 7 strong, 6 moderate, and 3 weak. We therefore call this comparator a \textbf{position-matched, content-similar other-instance donor}. The ratings indicate heterogeneous levels of receiver-visible relatedness.

\subsection{Aggregation and inference}
\label{sec:inference}

Within each story, we first average repeated measurements: presentation swaps and directed paths in both tasks, plus three symbolic codebooks in Moral Stories. Target-story aggregates are the resampling units. Moral Stories uses 10,000 shared cluster-bootstrap draws with simultaneous intervals over its four prespecified contrasts (Table~\ref{tab:moral-primary-family}); the direct Qwen-aligned current-minus-nearest contrast is post-outcome secondary. The original intervals condition on the donor pool and assignment and do not propagate donor-role dependence. Appendix~\ref{app:dependency-update} supplements them with connected-component resampling and node-deletion sensitivity.

\mbox{XStoryCloze} specifies three primary contrasts before outcome inspection: \mbox{HET-3} current minus donor, Qwen-aligned current minus donor, and their paired interaction. We average six paths and two presentation swaps inside each of 96 stories, then use 10,000 shared story-cluster draws and a standardized maximum-absolute-deviation critical value. The reported 95\% intervals are simultaneous over all three contrasts and conditional on the disjoint one-to-one donor assignment.

\section{Experimental Design}
\label{sec:design}

\subsection{Fixed systems and receiver interfaces}

The Moral Stories development phase fixed the model roster and receiver interface before the cross-task execution. Five checkpoints form two overlapping alignment contexts. The heterogeneous three-checkpoint system (\mbox{HET-3}) spans Gemma, Llama, and Qwen lineages. The Qwen-lineage alignment system (Qwen-aligned) groups two Qwen checkpoints with a Qwen-backbone distillation. Qwen3-8B provides a shared anchor. Early and late views follow one-third and two-thirds depth rules (floored for nonmultiples of three). The same phase set rank 2, the receiver target-norm rule, and patch coefficient \(\alpha=0.5\). \mbox{XStoryCloze} refits task-specific alignment states and target norms while retaining the roster, layer rules, rank, and patch coefficient. The checkpoints follow the Gemma 3, Llama 3, Qwen3, and DeepSeek-R1 releases \citep{gemmateam2025gemma3,grattafiori2024llama3,yang2025qwen3,deepseekai2025deepseekr1}.

Table~\ref{tab:systems} lists the checkpoints and receiver layers used in both task environments.

\begin{table*}[!t]
\centering
\small
\setlength{\tabcolsep}{4pt}
\begin{tabular}{@{}p{0.28\linewidth}p{0.34\linewidth}rr@{}}
\toprule
Model & System & Early / late layer & Hidden size \\
\midrule
Gemma-3-12B-IT & \mbox{HET-3} & 16 / 32 & 3,840 \\
Llama-3.2-3B-Instruct & \mbox{HET-3} & 9 / 18 & 3,072 \\
Qwen3-8B & Both & 12 / 24 & 4,096 \\
DeepSeek-R1-Distill-Qwen-7B & Qwen-aligned & 9 / 18 & 3,584 \\
Qwen3-14B & Qwen-aligned & 13 / 26 & 5,120 \\
\bottomrule
\end{tabular}
\caption{Fixed checkpoints and one-based transformer-block indices. Each system contributes all six directed non-self paths. Qwen3 thinking mode is disabled. ``System'' denotes a fitted alignment context; Qwen3-8B appears in both with identical weights and receiver settings.}
\label{tab:systems}
\end{table*}

\subsection{Two task environments}

Moral Stories provides structured pairs of norm-consistent and norm-violating actions \citep{emelin2021moralstories}. Alignment uses 512 training pairs; the controlled evaluation uses 64 previously unused training pairs. The receiver sees the norm, situation, intention, and selected action; the renderer omits the consequence. Three counterbalanced symbolic codebooks define the forced readout.

\mbox{XStoryCloze} asks which of two candidate endings coherently completes a four-sentence story \citep{lin2022xstorycloze}. We use the English split of \texttt{juletxara/xstory\_cloze} at revision prefix \texttt{12deb81b}; the artifact records the full revision and SHA-256 digests of both English TSV splits. Formation uses 256 training stories, 64 additional stories support interface checks, and confirmation uses 96 balanced evaluation-split targets. Candidate order is evaluated in canonical and swapped presentations with teacher-forced scoring of the A/B label.

\subsection{Outcome-access chronology and reproducibility}

The Moral Stories \mbox{HET-3} donor tests were specified after a 28-pair development study and before the 64-pair outcomes. The symmetric Qwen-aligned current-minus-nearest estimate and the \mbox{HET-3}-minus-Qwen-aligned interaction are secondary. For \mbox{XStoryCloze}, the symmetric primary family, data roles, one-to-one assignment, numerical decision criteria, and analysis code were fixed internally before the reported receiver outcomes were accessed. Receiver-stratified and robust-location analyses were added after release. The dependence, shuffled-donor, and rank/strength comparisons in Appendix~\ref{app:reviewer-comparisons} are reviewer-requested, post-outcome sensitivity analyses; they do not replace either primary family.

The release\footnote{\url{https://github.com/CaoGongce02/current-vector-selectivity}} supports released-row statistical reconstruction for Moral Stories and an end-to-end \mbox{XStoryCloze} chain from role construction and donor assignment through model execution and inference. One offline command verifies both packages and reconstructs the original statistics; supplementary comparisons have separate analysis and verification scripts. Full \mbox{XStoryCloze} execution requires the five upstream checkpoints under their licenses.

\subsection{Receiver runs and endpoints}

Every receiver run retains the current target text. Moral Stories evaluates exact zero, a cross-fitted generic answer direction, the current vector, the matcher-nearest donor, and a shuffled same-answer donor. \mbox{XStoryCloze} evaluates exact zero, a generic direction, the current vector, and the assigned position-matched donor. The current and donor arms use the same formation-derived receiver norm and coefficient. The zero and generic arms measure intervention efficacy; complete-vector substitution tests vector identity.

The primary endpoint is the teacher-forced gold margin. It is a log-odds contrast, so a current-minus-donor value of \(z\) corresponds to an odds multiplier of \(e^z\) within that task and interface. Forced-choice accuracy is computed from the sign of the same margin and reported separately because a larger margin need not cross the decision boundary. The \mbox{XStoryCloze} release contains 2,304 story-system-path-presentation rows and 9,216 logical arm cells. Moral Stories additionally averages three counterbalanced codebooks. Greedy-generation schedules are secondary diagnostics and do not enter either primary family.

\section{Results}
\label{sec:results}

\begin{figure*}[t]
\centering
\includegraphics[width=\textwidth]{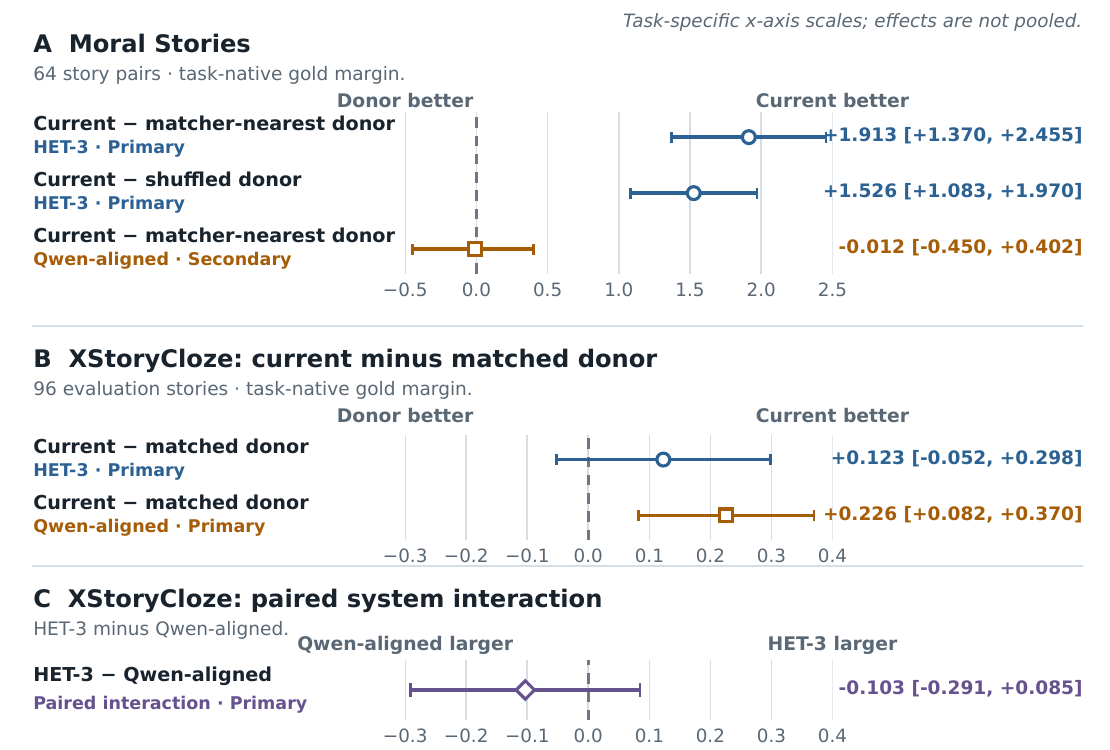}
\caption{Current-minus-donor evidence across two task environments. Points are task-native gold margin effects. Panel A uses the Moral Stories scale; Panels B--C share the \mbox{XStoryCloze} scale, and the two task scales are not pooled. Moral Stories intervals are conditional on the fixed donor assignment; the \mbox{HET-3} rows belong to the prespecified simultaneous family and the Qwen-aligned row is secondary. \mbox{XStoryCloze} intervals are simultaneous over its three prespecified contrasts. Donor/current direction labels apply to Panels A--B; Panel C reports the \mbox{HET-3} minus Qwen-aligned interaction. Accuracy changes are reported separately.}
\label{fig:cross-task}
\end{figure*}

Figure~\ref{fig:cross-task} summarizes the task-level contrasts. The two tasks use different native margin scales and are interpreted separately.

\subsection{Moral Stories: \mbox{HET-3} current vectors exceed matched donors}
\label{sec:moral-results}

\mbox{HET-3} current vectors exceed matcher-nearest donors by +1.913 [1.370, 2.455] gold margin units and shuffled donors by +1.526 [1.083, 1.970]. All 18 path-by-codebook cells have positive point estimates. The effect changes decisions: current-vector accuracy is 93.53\%, compared with 84.90\% for matcher-nearest donors, 87.46\% for shuffled donors, and 85.85\% at exact zero.

Because the same 64 stories also supply donors, we additionally analyze their dependency graph. The nearest-donor graph has 18 components (largest 25 stories), while the shuffled graph is connected. Resampling whole nearest-donor components gives a margin interval of [1.405, 2.569] and an accuracy difference of +8.64 percentage points [5.83, 12.59]. These are marginal 95\% sensitivity intervals under component independence, not replacements for the original simultaneous family. Node deletion also retains positive means for both comparators. Appendix~\ref{app:dependency-update} gives the assumptions and alternative estimates; none of these analyses resamples donor construction.

The Qwen-aligned system shows strong oracle generic steering, +1.978 [1.482, 2.473] versus zero, while the prespecified current-minus-generic contrast is \ensuremath{-}0.226 [\ensuremath{-}0.639, 0.188]. The direct current-minus-nearest comparison, specified after outcome inspection, is \ensuremath{-}0.012 [\ensuremath{-}0.450, 0.402]. The generic comparison supports an intervention effect; the direct current-minus-nearest interval includes zero.

\subsection{\mbox{XStoryCloze}: a smaller effect in continuous margin}

The prespecified \mbox{XStoryCloze} family yields a different pattern. Qwen-aligned current vectors exceed position-matched, content-similar donors by +0.226 [0.082, 0.370], a standardized effect of 0.373. Because the binary gold margin is a log-odds contrast, the mean corresponds to a geometric mean gold-odds multiplier of 1.254 [1.086, 1.448] for current versus donor. \mbox{HET-3} is +0.123 [\ensuremath{-}0.052, 0.298]. The paired \mbox{HET-3}-minus-Qwen-aligned interaction is \ensuremath{-}0.103 [\ensuremath{-}0.291, 0.085]; the task supports the Qwen-aligned contrast without establishing a system ordering.

Table~\ref{tab:cross-task} places the task-specific contrasts and their evidence status side by side without pooling their scales.

\begin{table*}[!t]
\centering
\small
\setlength{\tabcolsep}{4pt}
\begin{tabular}{@{}p{0.25\linewidth}p{0.29\linewidth}rrr@{}}
\toprule
Task and evidence status & System / contrast & Effect & 95\% CI & Stories \\
\midrule
Moral Stories (primary) & \mbox{HET-3} current \ensuremath{-} nearest donor & 1.913 & [1.370, 2.455] & 64 \\
Moral Stories (primary) & \mbox{HET-3} current \ensuremath{-} shuffled donor & 1.526 & [1.083, 1.970] & 64 \\
Moral Stories (secondary) & Qwen-aligned current \ensuremath{-} nearest donor & \ensuremath{-}0.012 & [\ensuremath{-}0.450, 0.402] & 64 \\
\mbox{XStoryCloze} (primary) & \mbox{HET-3} current \ensuremath{-} matched donor & 0.123 & [\ensuremath{-}0.052, 0.298] & 96 \\
\mbox{XStoryCloze} (primary) & Qwen-aligned current \ensuremath{-} matched donor & 0.226 & [0.082, 0.370] & 96 \\
\mbox{XStoryCloze} (primary) & \mbox{HET-3} \ensuremath{-} Qwen-aligned interaction & \ensuremath{-}0.103 & [\ensuremath{-}0.291, 0.085] & 96 \\
\bottomrule
\end{tabular}
\caption{Cross-task gold margin contrasts. Positive current-minus-donor effects favor the current vector; the interaction instead compares system effects. Moral Stories and \mbox{XStoryCloze} use different scales and are not pooled. The Moral Stories \mbox{HET-3} intervals belong to its four-contrast primary family; the Qwen-aligned interval is post-outcome secondary. The three \mbox{XStoryCloze} intervals form one simultaneous family. Accuracy is reported in Sections~\ref{sec:moral-results} and \ref{sec:behavior-results}.}
\label{tab:cross-task}
\end{table*}

\subsection{Selectivity, intervention efficacy, and accuracy separate}
\label{sec:behavior-results}

Accuracy remains near ceiling. \mbox{HET-3} current and donor accuracy are 96.61\% and 96.70\%; Qwen-aligned current and donor accuracy are 95.14\% and 95.75\%. In the 1,152 Qwen-aligned story--path--presentation cells, current alone is correct in 2 and donor alone in 9. The \mbox{XStoryCloze} result is therefore a graded preference effect, not an accuracy gain. Appendix~\ref{app:complete-results} reports presentation, path, receiver, robust-location, leave-one-model-out, and donor-similarity analyses.

The four arms' mean gold margins distinguish the current-vector effect from the donor's effect. For \mbox{HET-3}, exact zero, generic, current, and donor means are 11.263, 10.788, 11.227, and 11.104. For Qwen-aligned, they are 14.780, 14.892, 14.993, and 14.767. In the simultaneous secondary family, Qwen-aligned current-minus-zero is +0.213 [0.107, 0.319], whereas donor-minus-zero is \ensuremath{-}0.013 [\ensuremath{-}0.160, 0.134]. Thus, the current vector increases the mean margin relative to both zero and the matched donor, without improving accuracy.

\subsection{Heterogeneity and execution precision}

Presentation-stratified estimates are +0.369 [0.082, 0.655] for canonical order and +0.083 [\ensuremath{-}0.087, 0.254] after swapping in Qwen-aligned. The canonical-minus-swapped intervals include zero for both systems. Simultaneous path-level intervals are positive for 3 of 12 paths, while all six receiver-specific intervals include zero. The shared-Qwen3-8B alignment-context contrast is +0.035 [\ensuremath{-}0.178, 0.249].

The Qwen-aligned story-level mean, median, 10\%-trimmed mean, and positive-story fraction are 0.226, 0.042, 0.139, and 60.4\%. Its point estimate remains positive after excluding any sender (0.153--0.266) or any receiver (0.175--0.328), and ranges from 0.149 to 0.263 across donor-similarity quartiles.

The executed \mbox{XStoryCloze} models use bfloat16 weights; logits are converted to FP32 after the forward pass. All 2,304 swap-level current-minus-donor values lie on a 0.0625 grid, and the 1,152 swap-averaged story-path values lie on the corresponding 0.03125 grid. The primary interval describes the mean effect under this bfloat16 execution. Story-level variation is substantially wider than one grid step. The robust-location summaries show a right-skewed distribution, rather than a uniform effect across stories.

Appendix~\ref{app:shuffled-update} tests an additional same-label shuffled assignment. Appendix~\ref{app:rank-strength} reports all five added rank/strength configurations under a separate 15-contrast simultaneous family, retaining the original primary comparison at rank 2 and \(\alpha=0.5\).

\subsection{Results vary across task--interface configurations}

The same checkpoint roster and layer rules yield a strong \mbox{HET-3} result on Moral Stories and a smaller Qwen-aligned margin result on \mbox{XStoryCloze}. Neither the paired \mbox{XStoryCloze} system interaction nor the shared Qwen3-8B alignment-context contrast excludes zero, leaving no stable ordering of the two alignment systems across tasks.

An answer-axis intervention on the 28-pair Moral Stories development sample probes one source of this variation. Injecting matched answer-axis occupancy into \mbox{HET-3} changes selectivity by \ensuremath{-}1.247 [\ensuremath{-}1.764, \ensuremath{-}0.730]. Removing the axis from Qwen-aligned changes selectivity by +0.301 [\ensuremath{-}0.551, 1.154], and the matched-occupancy \mbox{HET-3}-minus-Qwen-aligned gap remains +1.450 [0.944, 1.956]. The injection supports a suppressive effect under this manipulation without explaining the full gap. Joint five-model alignment, pairwise alignment, alternative extraction positions, and answer-tail subtraction likewise leave the gap open (Appendix~\ref{app:mechanism-diagnostics}).

Generation provides a separate endpoint. Boundary-only, first-four-token, and every-token schedules were compared under greedy decoding. Persistent patching improves current-versus-random accuracy while preserving parse coverage, whereas the first-four current-versus-nearest comparison remains imprecise. The matched donor result is thus strongest at the forced task-native readout, and persistent injection does not convert it into a confirmed donor-specific generation gain.

\section{Discussion}
\label{sec:discussion}

\subsection{What the audit identifies}

Matched substitution defines a functional compatibility contrast. We compare the current vector with another complete, label-oriented vector at the same receiver interface and norm. Moral Stories preserves a semantic binary answer class; \mbox{XStoryCloze} preserves the source candidate-position label and adds receiver-visible content matching. A positive contrast means that the current vector produces a larger receiver-side margin than the assigned donor after the common orientation and intervention mechanics have been fixed.

Decision-relevant content, instance-specific geometric compatibility, and local calibration can all produce this contrast. The estimand measures the functional value lost when the complete current vector is replaced, but does not distinguish these explanations. We use \emph{current-vector selectivity} for this receiver-side property.

\subsection{Evaluation implications}

Efficacy and selectivity require different baselines, and neither substitutes for evaluating the downstream endpoint. Moral Stories combines a positive current-minus-donor margin with decision changes. \mbox{XStoryCloze} combines positive Qwen-aligned current-minus-zero and current-minus-donor margins with saturated accuracy. The same comparison can be used for residual patches, latent prefixes, KV transfer, or learned inter-agent channels whenever a complete other-instance message can enter the receiver through the same interface.

This ordering also clarifies how to compare latent interfaces. Zero and generic controls establish whether an interface can move the readout. Matched other-instance substitution then tests whether vector identity matters after orientation, norm, and receiver mechanics are fixed. The downstream metric determines whether that selectivity changes a decision, calibration, or generated behavior. Reporting them separately avoids conflating steering with instance-specific transfer, or margin changes with task-performance gains.

The donor construction is part of the estimand rather than a preprocessing detail. A content matcher changes which alternative vector defines the counterfactual, while one-to-one assignment changes how donor evidence is shared across targets. Reports should name the donor rule and separate target-story variation from assignment variation. This study reports fixed-assignment intervals together with deletion, similarity-stratified, and leave-one-model sensitivities; a new donor construction defines a different comparison, not a wider interval.

\subsection{A reporting contract for residual transfer}

A residual-transfer report should state the answer orientation, receiver interface, norm rule, donor construction, and uncertainty unit. These choices define the comparison: a different orientation tests label steering, an unmatched norm tests intervention strength, and repeated donors change the dependence structure. Current-minus-zero, current-minus-donor, and the downstream metric should then be reported separately. Moral Stories and \mbox{XStoryCloze} show why: margin selectivity can coincide with decision changes or remain confined to a graded readout. This contract makes transfer results comparable without assuming that aligned residual coordinates form a natural shared language.

Together, these quantities distinguish three evidence profiles: interface efficacy without vector-identity selectivity, graded selectivity without a decision change, and decision-level selectivity. The same profile can compare residual patches, latent prefixes, KV states, and learned continuous channels.

\section{Conclusion}

At an oracle-oriented receiver interface, equal-norm donor substitution makes vector identity the changed factor. Relative to the assigned donors, Moral Stories shows a large \mbox{HET-3} margin effect accompanied by decision changes; \mbox{XStoryCloze} shows a smaller Qwen-aligned margin effect without an accuracy gain. The contrast separates current-vector effects that alter decisions from those confined to graded margins.

\phantomsection\label{end:main-content}

\section*{Limitations}

The primary experiments cover two English binary-choice tasks, five fixed checkpoints, rank-2 MAXVAR-GCCA, one layer pair per model, and patch strength \(\alpha=0.5\). The supplementary comparisons vary rank and strength on the existing \mbox{XStoryCloze} roles; they do not sample new tasks, models, or training runs. Moral Stories uses unused training-split pairs and symbolic readouts; \mbox{XStoryCloze} uses evaluation-split targets from a versioned mirror and a near-ceiling A/B readout.

Each estimate conditions on its donor construction. Moral Stories reuses the 64 stories as targets and donors under a TF-IDF assignment. Component-level inference additionally assumes independence between disconnected components; it does not propagate the global matching procedure. \mbox{XStoryCloze} uses disjoint banks, with a receiver-visible matched assignment and one supplementary same-label shuffle. These comparisons do not cover donor-bank or matcher variation, and neither task supplies same-story minimal counterfactuals.

Official labels orient both vectors, and the receiver sees the complete task input, making the estimand an oracle-conditioned compatibility measure. It can reflect decision-relevant content, geometric compatibility, or calibration; the single-token injection and teacher-forced margin do not test open-ended communication.

\mbox{XStoryCloze} used bfloat16 (BF16) model weights and converted logits to 32-bit floating point (FP32) after the forward pass. Appendix~\ref{app:numerical-precision} documents the resulting numerical grid.

\section*{Ethical considerations}

The study uses released checkpoints and established language datasets. Moral Stories labels reflect cultural and annotation assumptions and do not measure general moral understanding; \mbox{XStoryCloze} inherits its source dataset's collection and licensing conditions. The artifact contains public row identifiers, donor assignments, and model outcomes but no user data, model weights, or private benchmark payloads. Checkpoint weights remain under upstream licenses. Compute and checkpoint records support resource and reproducibility assessment.

\section*{Acknowledgments}

This study was supported by the Beijing Natural Science Foundation (Grant No.~4252052) through the project ``Multilevel Safety Alignment and Evolving Systems for Ethical Value Learning in Superintelligence.'' It was also supported by the Innovation Pilot Program of Renmin University of China (Project No.~297526503509) through the project ``A Framework for Evaluating and Enhancing the Ethics and Safety of AI Agents.''

Generative AI tools were used for code modification, language polishing, and literature search.

\bibliography{references}

\begin{thebibliography}{29}
\providecommand{\natexlab}[1]{#1}

\bibitem[{Athanasiadis et~al.(2026)Athanasiadis, Karmush, and
  Felsberg}]{athanasiadis2026grounding}
Ioannis Athanasiadis, Anmar Karmush, and Michael Felsberg. 2026.
\newblock \href {https://openreview.net/forum?id=yUEFOa32CG} {Grounding
  functional similarity by invariance-aware model stitching}.
\newblock In \emph{Proceedings of the 43rd International Conference on Machine
  Learning}, volume 306 of \emph{Proceedings of Machine Learning Research},
  pages 4283--4304.

\bibitem[{Bansal et~al.(2021)Bansal, Nakkiran, and
  Barak}]{bansal2021revisiting}
Yamini Bansal, Preetum Nakkiran, and Boaz Barak. 2021.
\newblock \href
  {https://proceedings.neurips.cc/paper_files/paper/2021/hash/01ded4259d101feb739b06c399e9cd9c-Abstract.html}
  {Revisiting model stitching to compare neural representations}.
\newblock In \emph{Advances in Neural Information Processing Systems},
  volume~34, pages 225--236.

\bibitem[{Chen et~al.(2026)Chen, Zhang, Wu, Tremblay, Blukis, Birchfield,
  Vidal, Velasquez, Liu, and Qu}]{chen2026dense}
Siyi Chen, Xiaoyan Zhang, Meng Wu, Jonathan Tremblay, Valts Blukis, Stan
  Birchfield, Rene Vidal, Alvaro Velasquez, Sijia Liu, and Qing Qu. 2026.
\newblock \href {https://arxiv.org/abs/2606.13594} {See what {I} see, know what
  {I} think: Dense latent communication across heterogeneous agents}.
\newblock \emph{Preprint}, arXiv:2606.13594.

\bibitem[{{DeepSeek-AI}(2025)}]{deepseekai2025deepseekr1}
{DeepSeek-AI}. 2025.
\newblock \href {https://doi.org/10.1038/s41586-025-09422-z} {{DeepSeek-R1}
  incentivizes reasoning in {LLM}s through reinforcement learning}.
\newblock \emph{Nature}, 645:633--638.

\bibitem[{Du et~al.(2026)Du, Wang, Bai, Cao, Zhu, Cheng, Zheng, Chen, and
  Ying}]{du2026interlat}
Zhuoyun Du, Runze Wang, Huiyu Bai, Zouying Cao, Xiaoyong Zhu, Yu~Cheng,
  Bo~Zheng, Wei Chen, and Haochao Ying. 2026.
\newblock \href {https://doi.org/10.18653/v1/2026.acl-long.1248} {Enabling
  agents to communicate entirely in latent space}.
\newblock In \emph{Proceedings of the 64th Annual Meeting of the Association
  for Computational Linguistics}, pages 27106--27129.

\bibitem[{Emelin et~al.(2021)Emelin, Bras, Hwang, Forbes, and
  Choi}]{emelin2021moralstories}
Denis Emelin, Ronan~Le Bras, Jena~D. Hwang, Maxwell Forbes, and Yejin Choi.
  2021.
\newblock \href {https://doi.org/10.18653/v1/2021.emnlp-main.54} {{Moral
  Stories}: Situated reasoning about norms, intents, actions, and their
  consequences}.
\newblock In \emph{Proceedings of the 2021 Conference on Empirical Methods in
  Natural Language Processing}, pages 698--718.

\bibitem[{Geiger et~al.(2025)Geiger, Ibeling, Zur, Chaudhary, Chauhan, Huang,
  Arora, Wu, Goodman, Potts, and Icard}]{geiger2025causalabstraction}
Atticus Geiger, Duligur Ibeling, Amir Zur, Maheep Chaudhary, Sonakshi Chauhan,
  Jing Huang, Aryaman Arora, Zhengxuan Wu, Noah Goodman, Christopher Potts, and
  Thomas Icard. 2025.
\newblock \href {https://www.jmlr.org/papers/v26/23-0058.html} {Causal
  abstraction: A theoretical foundation for mechanistic interpretability}.
\newblock \emph{Journal of Machine Learning Research}, 26(83):1--64.

\bibitem[{{Gemma Team} et~al.(2025)}]{gemmateam2025gemma3}
{Gemma Team} et~al. 2025.
\newblock \href {https://arxiv.org/abs/2503.19786} {{Gemma 3} technical
  report}.
\newblock \emph{Preprint}, arXiv:2503.19786.

\bibitem[{Grattafiori et~al.(2024)}]{grattafiori2024llama3}
Aaron Grattafiori et~al. 2024.
\newblock \href {https://arxiv.org/abs/2407.21783} {The {Llama 3} herd of
  models}.
\newblock \emph{Preprint}, arXiv:2407.21783.

\bibitem[{Holtzman et~al.(2021)Holtzman, West, Shwartz, Choi, and
  Zettlemoyer}]{holtzman2021surfaceform}
Ari Holtzman, Peter West, Vered Shwartz, Yejin Choi, and Luke Zettlemoyer.
  2021.
\newblock \href {https://doi.org/10.18653/v1/2021.emnlp-main.564} {Surface form
  competition: Why the highest probability answer isn't always right}.
\newblock In \emph{Proceedings of the 2021 Conference on Empirical Methods in
  Natural Language Processing}, pages 7038--7051.

\bibitem[{Huang et~al.(2025)Huang, Huang, Feng, Lei, and
  Lv}]{huang2025crossmodel}
Youcheng Huang, Chen Huang, Duanyu Feng, Wenqiang Lei, and Jiancheng Lv. 2025.
\newblock \href {https://doi.org/10.18653/v1/2025.acl-long.185} {Cross-model
  transferability among large language models on the platonic representations
  of concepts}.
\newblock In \emph{Proceedings of the 63rd Annual Meeting of the Association
  for Computational Linguistics}, pages 3686--3704.

\bibitem[{Huh et~al.(2024)Huh, Cheung, Wang, and Isola}]{huh2024platonic}
Minyoung Huh, Brian Cheung, Tongzhou Wang, and Phillip Isola. 2024.
\newblock \href {https://proceedings.mlr.press/v235/huh24a.html} {Position: The
  platonic representation hypothesis}.
\newblock In \emph{Proceedings of the 41st International Conference on Machine
  Learning}, volume 235 of \emph{Proceedings of Machine Learning Research},
  pages 20617--20642.

\bibitem[{Jiang et~al.(2020)Jiang, Xu, Araki, and Neubig}]{jiang2020know}
Zhengbao Jiang, Frank~F. Xu, Jun Araki, and Graham Neubig. 2020.
\newblock \href {https://aclanthology.org/2020.tacl-1.28/} {How can we know
  what language models know?}
\newblock \emph{Transactions of the Association for Computational Linguistics},
  8:423--438.

\bibitem[{Kettenring(1971)}]{kettenring1971canonical}
Jon~R. Kettenring. 1971.
\newblock \href {https://doi.org/10.1093/biomet/58.3.433} {Canonical analysis
  of several sets of variables}.
\newblock \emph{Biometrika}, 58(3):433--451.

\bibitem[{Kornblith et~al.(2019)Kornblith, Norouzi, Lee, and
  Hinton}]{kornblith2019similarity}
Simon Kornblith, Mohammad Norouzi, Honglak Lee, and Geoffrey Hinton. 2019.
\newblock \href {https://proceedings.mlr.press/v97/kornblith19a.html}
  {Similarity of neural network representations revisited}.
\newblock In \emph{Proceedings of the 36th International Conference on Machine
  Learning}, volume~97 of \emph{Proceedings of Machine Learning Research},
  pages 3519--3529.

\bibitem[{Lee et~al.(2025)Lee, Padhi, Ramamurthy, Miehling, Dognin, Nagireddy,
  and Dhurandhar}]{lee2025cast}
Bruce~W. Lee, Inkit Padhi, Karthikeyan~Natesan Ramamurthy, Erik Miehling,
  Pierre Dognin, Manish Nagireddy, and Amit Dhurandhar. 2025.
\newblock \href {https://openreview.net/forum?id=Oi47wc10sm} {Programming
  refusal with conditional activation steering}.
\newblock In \emph{International Conference on Learning Representations}.

\bibitem[{Li and Shi(2026)}]{li2026latentbridge}
Bojie Li and Noah Shi. 2026.
\newblock \href {https://arxiv.org/abs/2606.24470} {The latent bridge: A
  continuous slow--fast channel for real-time game agents}.
\newblock \emph{Preprint}, arXiv:2606.24470.

\bibitem[{Li et~al.(2024)Li, Yan, Wang, Tang, Ren, Srinivasan, and
  Jin}]{li2024verbalizer}
Shiyang Li, Jun Yan, Hai Wang, Zheng Tang, Xiang Ren, Vijay Srinivasan, and
  Hongxia Jin. 2024.
\newblock \href {https://doi.org/10.18653/v1/2024.findings-naacl.233}
  {Instruction-following evaluation through verbalizer manipulation}.
\newblock In \emph{Findings of the Association for Computational Linguistics:
  NAACL 2024}, pages 3678--3692.

\bibitem[{Lin et~al.(2022)Lin, Mihaylov, Artetxe, Wang, Chen, Simig, Ott,
  Goyal, Bhosale, Du, Pasunuru, Shleifer, Koura, Chaudhary, O'Horo, Wang,
  Zettlemoyer, Kozareva, Diab, Stoyanov, and Li}]{lin2022xstorycloze}
Xi~Victoria Lin, Todor Mihaylov, Mikel Artetxe, Tianlu Wang, Shuohui Chen,
  Daniel Simig, Myle Ott, Naman Goyal, Shruti Bhosale, Jingfei Du, Ramakanth
  Pasunuru, Sam Shleifer, Punit~Singh Koura, Vishrav Chaudhary, Brian O'Horo,
  Jeff Wang, Luke Zettlemoyer, Zornitsa Kozareva, Mona Diab, Veselin Stoyanov,
  and Xian Li. 2022.
\newblock \href {https://doi.org/10.18653/v1/2022.emnlp-main.616} {Few-shot
  learning with multilingual generative language models}.
\newblock In \emph{Proceedings of the 2022 Conference on Empirical Methods in
  Natural Language Processing}, pages 9019--9052.

\bibitem[{Quirke(2026)}]{quirke2026ablation}
Philip Quirke. 2026.
\newblock \href {https://arxiv.org/abs/2606.08292} {Necessary, decodable and
  reversible, yet not transferable: A stress test for attention-head role
  claims}.
\newblock \emph{Preprint}, arXiv:2606.08292.

\bibitem[{Ramesh and Li(2025)}]{ramesh2025communicating}
Vignav Ramesh and Kenneth Li. 2025.
\newblock \href {https://proceedings.mlr.press/v267/ramesh25a.html}
  {Communicating activations between language model agents}.
\newblock In \emph{Proceedings of the 42nd International Conference on Machine
  Learning}, volume 267 of \emph{Proceedings of Machine Learning Research},
  pages 51094--51116.

\bibitem[{Rimsky et~al.(2024)Rimsky, Gabrieli, Schulz, Tong, Hubinger, and
  Turner}]{rimsky2024steering}
Nina Rimsky, Nick Gabrieli, Julian Schulz, Meg Tong, Evan Hubinger, and
  Alexander Turner. 2024.
\newblock \href {https://doi.org/10.18653/v1/2024.acl-long.828} {Steering
  {Llama 2} via contrastive activation addition}.
\newblock In \emph{Proceedings of the 62nd Annual Meeting of the Association
  for Computational Linguistics}, pages 15504--15522.

\bibitem[{Rodriguez et~al.(2025)Rodriguez, Blaas, Klein, Zappella, Apostoloff,
  Cuturi, and Suau}]{rodriguez2025activationtransport}
Pau Rodriguez, Arno Blaas, Michal Klein, Luca Zappella, Nicholas Apostoloff,
  Marco Cuturi, and Xavier Suau. 2025.
\newblock \href {https://openreview.net/forum?id=l2zFn6TIQi} {Controlling
  language and diffusion models by transporting activations}.
\newblock In \emph{International Conference on Learning Representations}.

\bibitem[{Smith et~al.(2025)Smith, Mannering, and Marcu}]{smith2025functional}
Damian Smith, Harvey Mannering, and Antonia Marcu. 2025.
\newblock \href {https://proceedings.mlr.press/v267/smith25a.html} {Functional
  alignment can mislead: Examining model stitching}.
\newblock In \emph{Proceedings of the 42nd International Conference on Machine
  Learning}, volume 267 of \emph{Proceedings of Machine Learning Research},
  pages 55972--55998.

\bibitem[{Turner et~al.(2023)Turner, Thiergart, Leech, Udell, Vazquez, Mini,
  and MacDiarmid}]{turner2023activationengineering}
Alexander~Matt Turner, Lisa Thiergart, Gavin Leech, David Udell, Juan~J.
  Vazquez, Ulisse Mini, and Monte MacDiarmid. 2023.
\newblock \href {https://arxiv.org/abs/2308.10248} {Steering language models
  with activation engineering}.
\newblock \emph{Preprint}, arXiv:2308.10248.

\bibitem[{Wenzel(2026)}]{wenzel2026latent}
Markus Wenzel. 2026.
\newblock \href {https://arxiv.org/abs/2607.14103} {Latent communication
  between language model agents: Channels, alignment, and the limits of text}.
\newblock \emph{Preprint}, arXiv:2607.14103.

\bibitem[{Yang et~al.(2025)}]{yang2025qwen3}
An~Yang et~al. 2025.
\newblock \href {https://arxiv.org/abs/2505.09388} {{Qwen3} technical report}.
\newblock \emph{Preprint}, arXiv:2505.09388.

\bibitem[{Zhang and Nanda(2024)}]{zhang2024activationpatching}
Fred Zhang and Neel Nanda. 2024.
\newblock \href
  {https://proceedings.iclr.cc/paper_files/paper/2024/hash/06a52a54c8ee03cd86771136bc91eb1f-Abstract-Conference.html}
  {Towards best practices of activation patching in language models: Metrics
  and methods}.
\newblock In \emph{International Conference on Learning Representations}.

\bibitem[{Zhang and Emu(2026)}]{zhang2026causalaudit}
Huixiang Zhang and Mahzabeen Emu. 2026.
\newblock \href {https://arxiv.org/abs/2607.26773} {Do latent channels actually
  communicate? {A} causal audit of latent multi-agent {LLM}}.
\newblock \emph{Preprint}, arXiv:2607.26773.

\end{thebibliography}
\clearpage
\appendix
\begin{center}
{\Large\bfseries Appendix}
\end{center}
\vspace{0.25em}
\section{Exact Transport and Substitution Algebra}
\label{app:algebra}

\subsection{Alignment state}

For each model view, formation differences are projected onto at most 64 PCA components, whitened with a relative eigenvalue floor of \(10^{-8}\), and passed to MAXVAR-GCCA. The shared rank is 2, encoder ridge relative strength is 0.1, and decoder ridge is 0. The two tasks fit independent alignment states; no Moral Stories state is reused on \mbox{XStoryCloze}.

Let \(U_s\) and \(\lambda_s\) denote sender PCA components and scales, \(E_s\) the shared encoder, and \(D_r\) the receiver decoder. The raw linear application is

\[
F_{s\rightarrow r}(\delta)=\left(\frac{\delta U_s}{\lambda_s}\right)E_sD_r.
\]

Application excludes the fitted formation mean. Thus \(F\) is linear, \(F(-\delta)=-F(\delta)\), and the antisymmetric operator in Section~\ref{sec:transport} satisfies \(A=F\). We check this identity and the effective patch under presentation swaps; the operator does not add a nonlinear transformation.

\subsection{Formation-derived target norms}
\label{app:target-norms}

Let \(\mathcal{F}\) be the formation role for the task and \(\ell_r\) the receiver's late layer. We set
\[
\tau_r=\operatorname{median}_{i\in\mathcal{F}}
\left\|h_r^{\ell_r}(x_i^B)-h_r^{\ell_r}(x_i^A)\right\|_2.
\]
This uses the receiver's own late-view differences, before cross-model transport, and no evaluation outcomes. Table~\ref{tab:target-norms} gives the values. Both complete vectors are independently scaled to \(\tau_r\), so their nominal injected norm is \(\alpha\tau_r\). Oracle orientation changes the sign, not the norm. In Moral Stories, the generic-only direction is the cross-fitted mean of oriented vectors and is not renormalized to this complete-vector norm.

\begin{table*}[t]
\centering
\small
\setlength{\tabcolsep}{6pt}
\begin{tabular}{@{}lrr@{}}
\toprule
Receiver & Moral Stories \(\tau_r\) & \mbox{XStoryCloze} \(\tau_r\) \\
\midrule
Gemma-3-12B-IT & 8232.9757 & 7154.2362 \\
Llama-3.2-3B-Instruct & 5.2164 & 7.8090 \\
Qwen3-8B & 61.8955 & 47.1170 \\
DeepSeek-R1-Distill-Qwen-7B & 12.5830 & 41.4995 \\
Qwen3-14B & 81.4747 & 49.2282 \\
\bottomrule
\end{tabular}
\caption{Formation-only receiver target norms, rounded to four decimal places. The primary patch coefficient is \(\alpha=0.5\). Norms are in each checkpoint's residual coordinates and are not comparable measures of intervention strength across checkpoints.}
\label{tab:target-norms}
\end{table*}

\subsection{Story attachment under presentation swap}

In \mbox{XStoryCloze}, candidate \(B\) minus candidate \(A\) defines the displayed difference. If candidate \(B\) is correct, the mapped direction receives orientation +1; if candidate \(A\) is correct, it receives \ensuremath{-}1. Swapping the candidate presentation negates the displayed difference and the answer orientation. Their product leaves the current story's oriented direction unchanged. Moral Stories uses the analogous blinded variant swap.

\subsection{Target-conditioned Moral Stories decomposition}

Seven folds are fixed before the 64-pair evaluation. For target \(i\), the oracle generic answer direction excludes its fold:

\[
\begin{aligned}
g_i&=\frac{1}{|\mathcal{T}_i|}\sum_{k\in\mathcal{T}_i}d_k,\\
\mathcal{T}_i&=\{k:y_k=y_i,\ f(k)\neq f(i)\}.
\end{aligned}
\]

Every donor residual is computed against the target axis, \(e_{j\mid i}=d_j-g_i\). The executed arms are

\[
\begin{aligned}
v_i^{\mathrm{current}}&=g_i+e_{i\mid i}=d_i,\\
v_i^{\mathrm{nearest}}&=g_i+e_{h(i)\mid i}=d_{h(i)},\\
v_i^{\mathrm{shuffle}}&=d_{s(i)}.
\end{aligned}
\]

This target conditioning closes the algebra even when target and donor occupy different cross-fitting folds.

\section{Data, Prompts, and Donor Construction}
\label{app:data}

\subsection{Moral Stories roles}

The formation role contains 512 training-split pairs. The controlled primary role contains 64 previously unused training pairs; a preceding 28-pair role supplied development evidence before the four-row primary family was frozen. Released pair rows and donor assignments identify every observation entering the statistical reconstruction. The receiver prompt renders the norm, situation, intention, and selected action, followed by one of three counterbalanced symbolic answer codebooks. The executed renderer removes the consequence field. Donor matching and receiver prompting therefore use different field sets.

\subsection{\mbox{XStoryCloze} roles and provenance}

The English mirror snapshot is \texttt{juletxara/xstory\_cloze} at revision prefix \texttt{12deb81b}, under CC BY-SA 4.0; the artifact records the full 40-character revision. It contains 360 training and 1,511 evaluation stories. Formation uses 256 training stories; 64 others support interface checks, and 40 remain unused. Confirmation uses 96 balanced evaluation targets. Matching alone decodes a separate bank of 256 evaluation stories; the assigned 96 donors are disjoint from targets, leaving 1,159 stories outside execution.

The original \mbox{XStoryCloze} paper reports 1,510 evaluation examples, whereas the published recipe and versioned mirror contain 1,511. The artifact records SHA-256 digests of the frozen English train and evaluation TSV files and does not alter either count.

\subsection{\mbox{XStoryCloze} renderer}

The receiver sees the four-sentence context and both candidate endings, then answers A or B. Canonical and swapped orders are rendered separately. The primary score is the teacher-forced log probability of the exact answer-label sequence. Qwen thinking mode is disabled; prompts longer than 512 tokens are rejected. For all five checkpoints, the artifact records SHA-256 digests of rendered-prompt bytes, chat templates, \texttt{tokenizer\_config.json}, and prompt/A/B token-ID sequences.

\subsection{Receiver-visible assignment audit}

Target and donor text comprises the context and both endings. A versioned \texttt{all-mpnet-base-v2} checkpoint embeds the rendered content. Eligible edges connect same-answer-label, different-story candidates with cosine similarity at least 0.25, character- and word-length ratios at most 1.75, trigram Jaccard similarity at most 0.80, and normalized edit similarity at most 0.92. Global one-to-one assignment minimizes \((1-\text{cosine})+0.35|\log r_{\mathrm{char}}|+0.20|\log r_{\mathrm{word}}|+0.10|\log r_{\mathrm{norm}}|\), subject to target--donor disjointness and no reuse. Deterministic row and column indices break ties. Outcome margins are absent from the assignment procedure.

For a positive quantity \(q\), the ratio is \(\max(q_i,q_j)/\min(q_i,q_j)\). Here \(r_{\mathrm{char}}\) and \(r_{\mathrm{word}}\) use character and word counts, and \(r_{\mathrm{norm}}\) uses the L2 norms of the text embeddings before unit normalization. The last quantity is a matcher feature, not a residual-vector or injected-patch norm.

Table~\ref{tab:donor-quality} summarizes the assignment geometry and the outcome-blind construct audit.

\begin{table}[htbp]
\centering
\small
\setlength{\tabcolsep}{4pt}
\begin{tabular}{@{}p{0.60\columnwidth}r@{}}
\toprule
Matcher quantity & Value \\
\midrule
Assignments / unique donors & 96 / 96 \\
Minimum cosine & 0.509 \\
First quartile & 0.615 \\
Median & 0.669 \\
Mean & 0.658 \\
Third quartile & 0.698 \\
Maximum & 0.844 \\
Outcome-blind single-rater audit: strong / moderate / weak & 7 / 6 / 3 \\
\bottomrule
\end{tabular}
\caption{\mbox{XStoryCloze} donor-quality audit. The outcome-blind single-rater review supports related but heterogeneous receiver-visible donors.}
\label{tab:donor-quality}
\end{table}

\section{Positioning Against the Closest Substitution Audits}
\label{app:closest-audits}

Table~\ref{tab:closest-audits} compares our design with prior substitution audits. Its final row summarizes the controls: equal-norm complete vectors rule out intervention magnitude, one-to-one donors prevent repeated-donor concentration, the fixed receiver interface removes hook and readout changes, and exact antisymmetry preserves story attachment under candidate-order swaps. These controls refine the current-versus-other-instance question for cross-model residual transfer; they do not claim priority for substitution auditing itself.

\begin{figure*}[t]
\centering
\includegraphics[width=\textwidth]{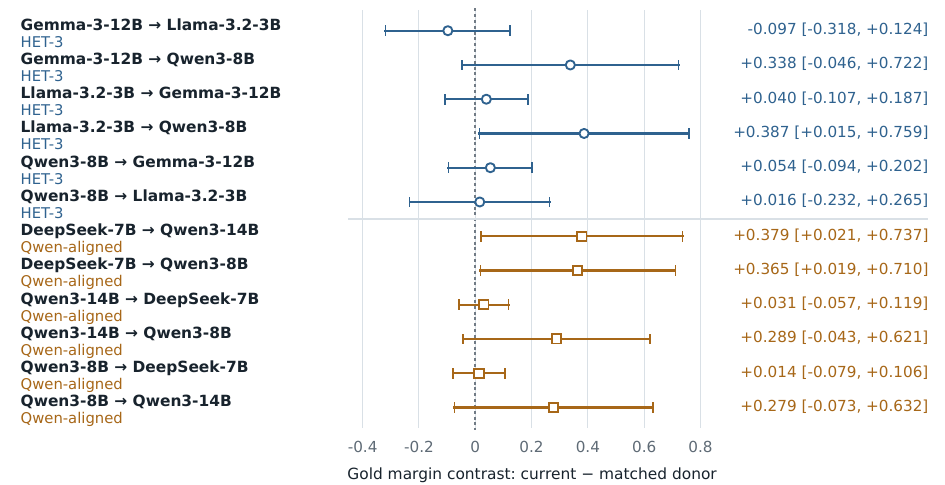}
\caption{Current-minus-donor gold margin effects for all 12 directed paths. Intervals are simultaneous over the path-level family. Three path intervals are above zero; nine include zero. Path labels abbreviate Gemma-3-12B-IT as Gemma-3-12B, Llama-3.2-3B-Instruct as Llama-3.2-3B, and DeepSeek-R1-Distill-Qwen-7B as DeepSeek-7B.}
\label{fig:path-forest}
\end{figure*}

\section{Complete Results and Robustness}
\label{app:complete-results}

Figure~\ref{fig:path-forest} and Table~\ref{tab:path-effects} through Table~\ref{tab:presentation} report the \mbox{XStoryCloze} path, receiver, outcome, and robustness analyses. Table~\ref{tab:moral-accuracy} and Table~\ref{tab:moral-overlap} report Moral Stories accuracy and donor-overlap sensitivity.

\subsection{\mbox{XStoryCloze} path, receiver, and outcome analyses}

The order-stratified estimates show how the aggregate current-minus-donor contrast changes across candidate presentations. In the released story--path--presentation cells, current-only versus donor-only correct counts are 0 versus 1 for \mbox{HET-3} and 2 versus 9 for Qwen-aligned. Table~\ref{tab:path-effects} gives all path estimates; Table~\ref{tab:receiver-sensitivity} aggregates by receiver. Table~\ref{tab:xsc-accuracy}, Table~\ref{tab:xsc-margins}, and Table~\ref{tab:xsc-secondary} report accuracy, absolute margins, and secondary arm contrasts, while Table~\ref{tab:xsc-robustness} and Table~\ref{tab:presentation} summarize robustness and presentation-order sensitivity.

\begin{table*}[tp]
\centering
\small
\setlength{\tabcolsep}{4pt}
\begin{tabular}{@{}p{0.28\linewidth}p{0.34\linewidth}rr@{}}
\toprule
System & Sender \ensuremath{\rightarrow} receiver & Effect & Simultaneous 95\% CI \\
\midrule
\mbox{HET-3} & Gemma-3-12B \ensuremath{\rightarrow} Llama-3.2-3B & \ensuremath{-}0.097 & [\ensuremath{-}0.318, 0.124] \\
\mbox{HET-3} & Gemma-3-12B \ensuremath{\rightarrow} Qwen3-8B & 0.338 & [\ensuremath{-}0.046, 0.722] \\
\mbox{HET-3} & Llama-3.2-3B \ensuremath{\rightarrow} Gemma-3-12B & 0.040 & [\ensuremath{-}0.107, 0.187] \\
\mbox{HET-3} & Llama-3.2-3B \ensuremath{\rightarrow} Qwen3-8B & \textbf{0.387} & \textbf{[0.015, 0.759]} \\
\mbox{HET-3} & Qwen3-8B \ensuremath{\rightarrow} Gemma-3-12B & 0.054 & [\ensuremath{-}0.094, 0.202] \\
\mbox{HET-3} & Qwen3-8B \ensuremath{\rightarrow} Llama-3.2-3B & 0.016 & [\ensuremath{-}0.232, 0.265] \\
Qwen-aligned & DeepSeek-7B \ensuremath{\rightarrow} Qwen3-14B & \textbf{0.379} & \textbf{[0.021, 0.737]} \\
Qwen-aligned & DeepSeek-7B \ensuremath{\rightarrow} Qwen3-8B & \textbf{0.365} & \textbf{[0.019, 0.710]} \\
Qwen-aligned & Qwen3-14B \ensuremath{\rightarrow} DeepSeek-7B & 0.031 & [\ensuremath{-}0.057, 0.119] \\
Qwen-aligned & Qwen3-14B \ensuremath{\rightarrow} Qwen3-8B & 0.289 & [\ensuremath{-}0.043, 0.621] \\
Qwen-aligned & Qwen3-8B \ensuremath{\rightarrow} DeepSeek-7B & 0.014 & [\ensuremath{-}0.079, 0.106] \\
Qwen-aligned & Qwen3-8B \ensuremath{\rightarrow} Qwen3-14B & 0.279 & [\ensuremath{-}0.073, 0.632] \\
\bottomrule
\end{tabular}
\caption{Path-level \mbox{XStoryCloze} effects over 96 stories. Aggregates in Table~\ref{tab:cross-task} weight the six paths equally within system. Bold marks intervals entirely above zero.}
\label{tab:path-effects}
\end{table*}

\begin{table*}[tp]
\centering
\small
\setlength{\tabcolsep}{4pt}
\begin{tabular}{@{}p{0.28\linewidth}p{0.34\linewidth}rr@{}}
\toprule
Alignment context & Receiver & Effect & Simultaneous 95\% CI \\
\midrule
\mbox{HET-3} & Gemma-3-12B-IT & 0.047 & [\ensuremath{-}0.091, 0.185] \\
\mbox{HET-3} & Llama-3.2-3B-Instruct & \ensuremath{-}0.040 & [\ensuremath{-}0.267, 0.186] \\
\mbox{HET-3} & Qwen3-8B & 0.362 & [\ensuremath{-}0.021, 0.745] \\
Qwen-aligned & DeepSeek-R1-Distill-Qwen-7B & 0.022 & [\ensuremath{-}0.066, 0.111] \\
Qwen-aligned & Qwen3-14B & 0.329 & [\ensuremath{-}0.013, 0.671] \\
Qwen-aligned & Qwen3-8B & 0.327 & [\ensuremath{-}0.012, 0.666] \\
Shared receiver & Qwen3-8B, \mbox{HET-3} \ensuremath{-} Qwen-aligned & 0.035 & [\ensuremath{-}0.178, 0.249] \\
\bottomrule
\end{tabular}
\caption{Post-outcome receiver-stratified sensitivity. The seven simultaneous intervals use 10,000 shared story-cluster draws over 96 stories. Receiver means average two incoming senders and both presentation orders. All six receiver-specific intervals and the shared-receiver contrast include zero; the point-estimate pattern is descriptive and does not replace the prespecified system-level family.}
\label{tab:receiver-sensitivity}
\end{table*}

\begin{table*}[tp]
\centering
\small
\setlength{\tabcolsep}{4pt}
\begin{tabular}{@{}llllll@{}}
\toprule
System & Exact zero & Generic accuracy & Current accuracy & Donor accuracy & Current \ensuremath{-} donor \\
\midrule
\mbox{HET-3} & 96.53\% & 96.70\% & 96.61\% & 96.70\% & \ensuremath{-}0.09 pp \\
Qwen-aligned & 95.83\% & 95.57\% & 95.14\% & 95.75\% & \ensuremath{-}0.61 pp \\
\bottomrule
\end{tabular}
\caption{\mbox{XStoryCloze} descriptive accuracy. Coverage is 100\% for every arm. These rows are secondary and unadjusted.}
\label{tab:xsc-accuracy}
\end{table*}

\begin{table*}[tp]
\centering
\small
\setlength{\tabcolsep}{4pt}
\begin{tabular}{@{}p{0.25\linewidth}p{0.29\linewidth}rrr@{}}
\toprule
System & Exact zero & Generic & Current & Matched donor \\
\midrule
\mbox{HET-3} & 11.263 & 10.788 & 11.227 & 11.104 \\
Qwen-aligned & 14.780 & 14.892 & 14.993 & 14.767 \\
\bottomrule
\end{tabular}
\caption{Mean \mbox{XStoryCloze} gold margins on the absolute four-arm scale. Intervals and machine-readable story aggregates are released in the artifact.}
\label{tab:xsc-margins}
\end{table*}

\begin{table*}[tp]
\centering
\small
\setlength{\tabcolsep}{4pt}
\begin{tabular}{@{}p{0.46\linewidth}p{0.31\linewidth}r@{}}
\toprule
Secondary contrast & Effect & Simultaneous 95\% CI \\
\midrule
\mbox{HET-3} current \ensuremath{-} zero & \ensuremath{-}0.036 & [\ensuremath{-}0.188, 0.116] \\
\mbox{HET-3} donor \ensuremath{-} zero & \ensuremath{-}0.159 & [\ensuremath{-}0.294, \ensuremath{-}0.023] \\
\mbox{HET-3} current \ensuremath{-} generic & \textbf{0.439} & \textbf{[0.256, 0.621]} \\
\mbox{HET-3} current \ensuremath{-} donor & 0.123 & [\ensuremath{-}0.074, 0.320] \\
Qwen-aligned current \ensuremath{-} zero & \textbf{0.213} & \textbf{[0.107, 0.319]} \\
Qwen-aligned donor \ensuremath{-} zero & \ensuremath{-}0.013 & [\ensuremath{-}0.160, 0.134] \\
Qwen-aligned current \ensuremath{-} generic & 0.101 & [\ensuremath{-}0.039, 0.241] \\
Qwen-aligned current \ensuremath{-} donor & \textbf{0.226} & \textbf{[0.064, 0.388]} \\
\bottomrule
\end{tabular}
\caption{Post-outcome eight-contrast simultaneous family, separate from the three primary contrasts. Bold marks intervals entirely above zero.}
\label{tab:xsc-secondary}
\end{table*}

\begin{table*}[tp]
\centering
\footnotesize
\setlength{\tabcolsep}{2.0pt}
\begin{tabular}{@{}p{0.17\linewidth}p{0.255\linewidth}p{0.255\linewidth}p{0.255\linewidth}@{}}
\toprule
System & Leave-one-sender range & Leave-one-receiver range & Cosine-quartile range \\
\midrule
\mbox{HET-3} & 0.078--0.167 & 0.003--0.205 & \ensuremath{-}0.024--0.234 \\
Qwen-aligned & 0.153--0.266 & 0.175--0.328 & 0.149--0.263 \\
\bottomrule
\end{tabular}
\caption{Descriptive robustness of current-minus-donor point estimates. The cosine-effect Pearson/Spearman correlations are \ensuremath{-}0.045/\ensuremath{-}0.093 for \mbox{HET-3} and \ensuremath{-}0.050/0.022 for Qwen-aligned.}
\label{tab:xsc-robustness}
\end{table*}

\begin{table*}[tp]
\centering
\footnotesize
\setlength{\tabcolsep}{2.5pt}
\begin{tabular}{@{}p{0.13\linewidth}p{0.25\linewidth}p{0.25\linewidth}p{0.25\linewidth}@{}}
\toprule
System & Canonical effect & Swapped effect & Canonical \ensuremath{-} swapped \\
\midrule
\mbox{HET-3} & 0.214 [\ensuremath{-}0.036, 0.463] & 0.032 [\ensuremath{-}0.206, 0.271] & 0.181 [\ensuremath{-}0.131, 0.494] \\
Qwen-aligned & \textbf{0.369 [0.082, 0.655]} & 0.083 [\ensuremath{-}0.087, 0.254] & 0.285 [\ensuremath{-}0.067, 0.638] \\
\bottomrule
\end{tabular}
\caption{Post-outcome presentation-stratified current-minus-donor effects. All six intervals are simultaneous within one shared-draw family over 96 stories. The primary estimand averages canonical and swapped presentations; this table measures heterogeneity around that aggregate. Bold marks the one order-specific interval entirely above zero.}
\label{tab:presentation}
\end{table*}

\subsection{Moral Stories behavior and donor overlap}
\label{app:moral-overlap}

Table~\ref{tab:moral-accuracy} reports arm-wise forced-symbolic accuracy. Table~\ref{tab:moral-overlap} summarizes target--donor graph structure and deletion sensitivity under the frozen assignments.

\begin{table*}[tp]
\centering
\small
\setlength{\tabcolsep}{3pt}
\begin{tabular*}{0.92\textwidth}{@{\extracolsep{\fill}}lrrrrr@{}}
\toprule
System & Exact zero & Generic & Current & Nearest donor & Shuffled donor \\
\midrule
\mbox{HET-3} & 85.85\% & 88.50\% & \textbf{93.53\%} & 84.90\% & 87.46\% \\
Qwen-aligned & 84.72\% & 88.02\% & 87.59\% & 87.80\% & \textbf{88.15\%} \\
\bottomrule
\end{tabular*}
\caption{Fixed-cell forced-symbolic accuracy on 64 Moral Stories pairs. Bold marks the highest accuracy within each system row. In \mbox{HET-3}, leave-one-direction current-minus-nearest estimates range from 1.693 to 2.122 and leave-one-checkpoint estimates from 1.347 to 2.372.}
\label{tab:moral-accuracy}
\end{table*}

\begin{table*}[tp]
\centering
\footnotesize
\setlength{\tabcolsep}{2.5pt}
\begin{tabular}{@{}p{0.17\linewidth}p{0.24\linewidth}p{0.54\linewidth}@{}}
\toprule
Donor arm & Assignment graph & Point-estimate sensitivity \\
\midrule
Matcher-nearest & 18 components; largest 25 & Mean 1.913; node deletion [1.823, 2.040]; component deletion [1.823, 1.997] \\
Shuffled & One connected component of 64 & Mean 1.526; node deletion [1.392, 1.605]; component deletion not estimable \\
\bottomrule
\end{tabular}
\caption{Moral Stories donor-overlap sensitivity. An edge links a target story to either variant-level donor. Single-node deletion removes every target effect in which the deleted story appears as target or donor. Component deletion is unavailable for the connected shuffled graph. Ranges are deterministic sensitivities, not replacement confidence intervals.}
\label{tab:moral-overlap}
\end{table*}

\section{Mechanism, Generation, and Statistical Details}
\label{app:mechanism}

\begin{table*}[t]
\centering
\small
\setlength{\tabcolsep}{5pt}
\begin{tabular}{@{}llrr@{}}
\toprule
System & Prespecified contrast & Gold margin & Simultaneous 95\% CI \\
\midrule
\mbox{HET-3} & Current \( - \) nearest donor & 1.913 & [1.370, 2.455] \\
\mbox{HET-3} & Current \( - \) shuffled donor & 1.526 & [1.083, 1.970] \\
Qwen-aligned & Generic \( - \) exact zero & 1.978 & [1.482, 2.473] \\
Qwen-aligned & Current \( - \) generic & \(-0.226\) & [\(-0.639\), 0.188] \\
\bottomrule
\end{tabular}
\caption{The complete four-contrast Moral Stories primary family on 64 story pairs. These original target-pair intervals condition on the fixed assignment. The Qwen-aligned generic comparisons have different estimands from the equal-norm current-minus-donor comparisons.}
\label{tab:moral-primary-family}
\end{table*}

\subsection{Mechanism diagnostics}
\label{app:mechanism-diagnostics}

The answer-axis study uses the 28-pair development sample, a rank-1 direction, and norm- and distance-matched shams. Removing the axis from Qwen-aligned changes selectivity by +0.301 [\ensuremath{-}0.551, 1.154]. Injecting matched occupancy into \mbox{HET-3} changes selectivity by \ensuremath{-}1.247 [\ensuremath{-}1.764, \ensuremath{-}0.730], while the matched-occupancy \mbox{HET-3}-minus-Qwen-aligned gap remains +1.450 [0.944, 1.956]. These estimates concern the development sample, not the later 64-pair evaluation. The injection supports a suppressive effect under this manipulation; the Qwen-aligned removal result and remaining gap do not establish a complete explanation.

Joint five-model alignment, pairwise alignment, answer-boundary extraction, content-end extraction, and answer-tail subtraction were also evaluated with matched shams. Their intervals do not close the Moral Stories system gap. These analyses are secondary diagnostics rather than evidence for a complete mechanism.

\subsection{Persistent generation}

Boundary-only, first-four-token, and every-token schedules were compared during greedy generation. Persistent patching improves current-versus-random accuracy while maintaining parse coverage, but the first-four current-versus-nearest comparison remains imprecise. The direct forced-readout donor comparison is therefore the principal endpoint.

\Needspace{8\baselineskip}
\subsection{Aggregation and simultaneous inference}

Each \mbox{XStoryCloze} row identifies a story, system, directed path, and presentation swap and contains scores for all arms. We compute current-minus-donor within each matched cell, then average the two swaps and six paths with equal weight within story. This gives one \mbox{HET-3} effect and one Qwen-aligned effect per story; their difference is the paired interaction. The result set contains 2,304 rows.

Let \(\bar\theta_k\) and \(\widehat{\mathrm{se}}_k\) denote the observed mean and bootstrap standard error for contrast \(k\in\{1,2,3\}\). Each draw resamples the same 96 story indices for all contrasts. The critical value is the 95th percentile of

\[
\max_k\left|\frac{\theta_k^{*(b)}-\bar\theta_k}{\widehat{\mathrm{se}}_k}\right|.
\]

With 10,000 draws and seed 2026080303, the resulting critical value is 2.3350. Intervals are \(\bar\theta_k\pm c\widehat{\mathrm{se}}_k\). Standardized effects divide the mean by the sample story-level standard deviation: 0.167 for \mbox{HET-3}, 0.373 for Qwen-aligned, and \ensuremath{-}0.130 for the interaction.

\subsection{Prespecified decision rule}

The prespecified positive decision requires a simultaneous lower bound above zero and a point estimate of at least 0.10. Qwen-aligned satisfies both; \mbox{HET-3} satisfies the point threshold but not the interval threshold. The interaction is reported separately from the two system-specific results. The same classifications hold at thresholds 0.05, 0.10, and 0.15: Qwen-aligned is positive, while \mbox{HET-3} and the interaction remain inconclusive. The artifact releases the outcome-blind planning simulation with its seed, draws, assumptions, critical values, and complete threshold neighborhood.

\subsection{Numerical precision audit}
\label{app:numerical-precision}

We load receiver weights in bfloat16 (BF16) and convert candidate logits to 32-bit floating point (FP32) after the forward pass, before log-probability scoring. All 2,304 swap-level current-minus-donor contrasts lie on a 0.0625 grid, and all 1,152 swap-averaged story--path contrasts lie on a 0.03125 grid. The artifact regenerates this diagnostic from released rows. The primary interval measures story variation under this BF16 execution, rather than agreement with higher-precision inference.

\subsection{Presentation-stratified secondary family}

For each story and system, current-minus-donor is averaged over six directed paths separately for \texttt{canonical\_ab} and \texttt{swapped\_ba}. The secondary family contains canonical, swapped, and canonical-minus-swapped contrasts for \mbox{HET-3} and Qwen-aligned. The same 10,000 story-index draws and seed 2026080303 are shared across all six contrasts; a maximum-absolute-deviation critical value of 2.5182 produces the simultaneous intervals in the presentation-order table in Appendix~\ref{app:complete-results}. This analysis was specified after release of the primary outcomes and does not replace the three-contrast prespecified family.

\subsection{Receiver-stratified secondary family}

For each receiver and story, the current-minus-donor contrast averages its two incoming senders and both presentation orders. Six receiver means and the paired shared-Qwen3-8B alignment-context difference form one post-outcome family. Ten thousand shared story-cluster draws with seed 2026080306 yield a standardized maximum-absolute-deviation critical value of 2.8084. The receiver-stratified table in Appendix~\ref{app:complete-results} reports the seven simultaneous intervals. System-level robust summaries use the same 96 story aggregates and report the mean, median, 10\% trimmed mean, and positive-story fraction.

\subsection{Conditional Moral Stories inference}

The original Moral Stories bootstrap resamples target-story effects after the donor assignment is fixed. Its intervals do not propagate donor-role dependence or rematching uncertainty. The dependency graph links target and donor roles from the released assignment ledger. Table~\ref{tab:moral-overlap} reports node- and component-deletion mean ranges; Appendix~\ref{app:dependency-update} adds dependence-aware sensitivity intervals under explicit assumptions. Table~\ref{tab:moral-primary-family} lists all four original primary contrasts together.

\FloatBarrier
\section{Reviewer-Requested Sensitivity Comparisons}
\label{app:reviewer-comparisons}

These analyses were added after the primary outcomes. They reuse the original evaluation roles and retain the reported primary estimates. Each analysis below specifies its own uncertainty calculation.

\subsection{Moral Stories donor-dependency analysis}
\label{app:dependency-update}

For each of the 64 stories, we average the current-minus-donor margin and accuracy differences over six paths, three codebooks, and two variants. We then form an undirected graph whose nodes are stories and whose edges connect every target to its assigned donor in either answer class. The nearest-donor graph has 18 connected components of sizes 25, 6, 3, and fifteen 2-story components. The shuffled graph contains one 64-story component.

For nearest donors, let component \(c\) contain \(n_c\) targets with effect sum \(S_c\). We draw 18 components with replacement and calculate the target-weighted mean
\[
\theta^*=\frac{\sum_{c\in\mathcal{C}^*}S_c}
{\sum_{c\in\mathcal{C}^*}n_c}.
\]
With 200,000 draws and seed 2026091701, percentile intervals are [1.405, 2.569] for the margin and [5.83, 12.59] percentage points for accuracy. The accuracy point estimate is 8.637 percentage points; the 8.63-point difference in the abstract subtracts the displayed, rounded arm accuracies. Table~\ref{tab:moral-dependency-update} also reports a cluster-robust CR1 interval with \(t_{17}\) critical values. Both calculations treat disconnected components as independent after conditioning on the assignment. The largest component contains 39.1\% of targets, so this is a small, unbalanced cluster analysis.

As a second sensitivity, we delete each story and every target effect that uses it as a donor. A delete-node jackknife uses the resulting 64 means and \(t_{63}\) critical values. This removes a story from both roles, but overlapping deletions and the fixed assignment limit its interpretation as a general sampling interval. Component resampling cannot estimate uncertainty for the connected shuffled graph. None of these procedures regenerates the donor pool or reruns matching.

\begin{table*}[t]
\centering
\small
\setlength{\tabcolsep}{5pt}
\begin{tabular}{@{}llrr@{}}
\toprule
Donor & Sensitivity calculation & Margin 95\% interval & Accuracy 95\% interval (pp) \\
\midrule
Nearest & Whole-component bootstrap & [1.405, 2.569] & [5.83, 12.59] \\
Nearest & Component CR1, \(t_{17}\) & [1.396, 2.429] & [5.62, 11.66] \\
Nearest & Delete-node jackknife, \(t_{63}\) & [1.048, 2.777] & [3.71, 13.56] \\
Shuffled & Delete-node jackknife, \(t_{63}\) & [0.933, 2.119] & [3.06, 9.09] \\
\bottomrule
\end{tabular}
\caption{Post-outcome \mbox{HET-3} donor-dependency sensitivities on Moral Stories. Nearest-donor point estimates are 1.913 margin units and 8.64 percentage points (pp); shuffled-donor estimates are 1.526 and 6.08 pp. Intervals are marginal, not a simultaneous family, and retain the stated fixed-assignment assumptions.}
\label{tab:moral-dependency-update}
\end{table*}

\subsection{An additional \mbox{XStoryCloze} shuffled assignment}
\label{app:shuffled-update}

We reuse the 96 targets and the original selected donor pool, permuting donors within source answer-position groups without reuse. This supplies one additional complete-vector comparator with the same orientation, norm, receiver prompt, and scoring rule. It changes assignment, not the embedding checkpoint or donor bank. Original-arm margins, prompt-token hashes, and patch norms agree with the released execution.

We average both presentations and six paths within story. The four margin contrasts in Table~\ref{tab:xsc-shuffled-update} share 200,000 story-bootstrap draws with seed 2026091702. A standardized maximum-absolute-deviation critical value of 2.4576 gives simultaneous 95\% intervals. Accuracy forms a separate four-contrast family, with critical value 2.3627. Qwen-aligned current-minus-shuffled margin is +0.208 [0.070, 0.347]. The matched-minus-shuffled intervals and all four accuracy intervals include zero. The Qwen-aligned mean margin contrast therefore remains positive under this additional same-label assignment.

\begin{table*}[t]
\centering
\small
\setlength{\tabcolsep}{4pt}
\begin{tabular}{@{}llrr@{}}
\toprule
System & Contrast & Margin effect [95\% CI] & Accuracy effect [95\% CI] (pp) \\
\midrule
\mbox{HET-3} & Current \( - \) shuffled & 0.123 [\(-0.060\), 0.306] & 0.52 [\(-0.24\), 1.28] \\
Qwen-aligned & Current \( - \) shuffled & 0.208 [0.070, 0.347] & \(-0.17\) [\(-1.30\), 0.95] \\
\mbox{HET-3} & Matched \( - \) shuffled & 0.000 [\(-0.182\), 0.182] & 0.61 [\(-0.33\), 1.54] \\
Qwen-aligned & Matched \( - \) shuffled & \(-0.018\) [\(-0.181\), 0.145] & 0.43 [\(-0.71\), 1.58] \\
\bottomrule
\end{tabular}
\caption{One additional same-label shuffled assignment on \mbox{XStoryCloze}. Margin and accuracy have separate four-contrast simultaneous families. ``Matched'' is the original position-matched, content-similar donor. The first matched-minus-shuffled margin rounds from 0.000217.}
\label{tab:xsc-shuffled-update}
\end{table*}

\subsection{Rank and intervention strength}
\label{app:rank-strength}

We evaluate five additional \mbox{XStoryCloze} configurations: ranks 4, 6, and 8 at \(\alpha=0.5\), and rank 2 at \(\alpha=0.25\) and 1.0. The checkpoints, extraction and injection layers, prompts, donor assignment, formation-derived target norms, and scoring rule remain unchanged. Higher-rank maps are fitted from the original 256-story formation activations. The primary rank-2, \(\alpha=0.5\) result retains its original three-contrast family and is not retested as part of this grid.

For each configuration, we first average six directed paths and both presentations within each of the 96 stories. The two system means and their paired difference yield 15 contrasts across five configurations. Ten thousand shared story-bootstrap draws, with seed 2026112601 and critical value 2.7407, give one simultaneous 95\% family (Table~\ref{tab:rank-strength}). These are post-outcome comparisons on the original evaluation roles.

Qwen-aligned current-minus-donor margin is +0.203 [0.014, 0.391] at rank 4. At ranks 6 and 8, the means remain positive at +0.165 [\(-0.010\), 0.340] and +0.132 [\(-0.051\), 0.316]; both intervals include zero. At rank 2, the effects are +0.109 [0.022, 0.197] for \(\alpha=0.25\) and +0.405 [0.108, 0.702] for \(\alpha=1.0\). All five \mbox{HET-3} intervals and all five paired system-difference intervals include zero. Accuracy differences are reported separately in Table~\ref{tab:rank-strength-accuracy}. These comparisons extend the positive Qwen-aligned margin finding to rank 4 and two additional intervention strengths.

\begin{table*}[t]
\centering
\small
\setlength{\tabcolsep}{8pt}
\begin{tabular}{@{}rrlrr@{}}
\toprule
Rank & \(\alpha\) & Contrast & Gold margin & Simultaneous 95\% CI \\
\midrule
4 & 0.50 & \mbox{HET-3} current \( - \) donor & 0.065 & [\(-0.086\), 0.216] \\
4 & 0.50 & Qwen-aligned current \( - \) donor & 0.203 & [0.014, 0.391] \\
4 & 0.50 & \mbox{HET-3} \( - \) Qwen-aligned & \(-0.138\) & [\(-0.320\), 0.044] \\
\addlinespace
6 & 0.50 & \mbox{HET-3} current \( - \) donor & 0.047 & [\(-0.114\), 0.209] \\
6 & 0.50 & Qwen-aligned current \( - \) donor & 0.165 & [\(-0.010\), 0.340] \\
6 & 0.50 & \mbox{HET-3} \( - \) Qwen-aligned & \(-0.118\) & [\(-0.289\), 0.053] \\
\addlinespace
8 & 0.50 & \mbox{HET-3} current \( - \) donor & \(-0.004\) & [\(-0.178\), 0.170] \\
8 & 0.50 & Qwen-aligned current \( - \) donor & 0.132 & [\(-0.051\), 0.316] \\
8 & 0.50 & \mbox{HET-3} \( - \) Qwen-aligned & \(-0.136\) & [\(-0.303\), 0.030] \\
\addlinespace
2 & 0.25 & \mbox{HET-3} current \( - \) donor & 0.066 & [\(-0.040\), 0.172] \\
2 & 0.25 & Qwen-aligned current \( - \) donor & 0.109 & [0.022, 0.197] \\
2 & 0.25 & \mbox{HET-3} \( - \) Qwen-aligned & \(-0.044\) & [\(-0.160\), 0.073] \\
\addlinespace
2 & 1.00 & \mbox{HET-3} current \( - \) donor & 0.204 & [\(-0.160\), 0.569] \\
2 & 1.00 & Qwen-aligned current \( - \) donor & 0.405 & [0.108, 0.702] \\
2 & 1.00 & \mbox{HET-3} \( - \) Qwen-aligned & \(-0.200\) & [\(-0.609\), 0.208] \\
\bottomrule
\end{tabular}
\caption{All five added rank/strength configurations. Each system contrast averages 96 stories; paired differences use the same stories. The 15 intervals form one simultaneous family, separate from the primary rank-2, \(\alpha=0.5\) analysis.}
\label{tab:rank-strength}
\end{table*}

\begin{table}[htbp]
\centering
\small
\setlength{\tabcolsep}{6pt}
\begin{tabular}{@{}rrrr@{}}
\toprule
Rank & \(\alpha\) & \mbox{HET-3} (pp) & Qwen-aligned (pp) \\
\midrule
4 & 0.50 & \(-0.17\) & \(-0.26\) \\
6 & 0.50 & 0.00 & \(-0.17\) \\
8 & 0.50 & 0.26 & \(-0.09\) \\
2 & 0.25 & \(-0.09\) & \(-0.52\) \\
2 & 1.00 & 0.17 & 0.35 \\
\bottomrule
\end{tabular}
\caption{Descriptive current-minus-donor accuracy differences for the added configurations, in percentage points. Values are story-averaged point estimates.}
\label{tab:rank-strength-accuracy}
\end{table}

\section{A Worked Ethical-Choice Example}
\label{app:worked-example}

Consider a Moral Stories case in which Chad has failed a mathematics test. He wants to attend a friend's birthday party without being grounded. When his parents ask about his grade, the paired actions contrast admitting the result with claiming he did well. The stated norm prohibits lying to family members. We use the norm-consistent action to illustrate the executed comparison.

The receiver reads Chad's norm, situation, intention, and selected action in both runs. Its assigned same-answer donor concerns Hank wiping his shoes before entering a friend's house. This donor matches the compliance label, not the ethical topic. The receiver never receives Hank's text. Each complete vector has its own official-label orientation and is rescaled to the same receiver norm.

We use Llama-3.2-3B-Instruct as sender and Gemma-3-12B-IT as receiver in \mbox{HET-3}. Under the 1/2 codebook, the correct symbol is 2. The current-vector run assigns pre-softmax logits 33.000 and 26.625 to symbols 2 and 1, giving a gold margin of 6.375. The donor run assigns logits 36.750 and 35.750, giving a margin of 1.000. Their difference is therefore
\[
\Delta=6.375-1.000=5.375.
\]
Both classifications are correct. The larger current-vector margin comes from the difference between correct and incorrect scores, even though its correct-symbol score alone is lower. This is a local margin contrast, not an accuracy improvement. The example was selected to explain the procedure; the reported task estimates average all paths, codebooks, variants, and stories.

\section{Reproducibility Package}
\label{app:reproducibility}

Unpack Software and Data into sibling directories. In Software, run \texttt{python} \texttt{verify\_with\_data.py}, setting \verb|--data-root| to the Data directory. The wrapper checks both manifests and combines the payload in a temporary directory. Its internal verifier checks hashes, code compilation, roles, assignments, and reconstruction of the original tables. Offline verification does not load model weights or execute new forwards.

For \mbox{XStoryCloze}, the package records data and checkpoint revisions, the role-selection seed, public row IDs, the exact donor assignment, and prompt/tokenization specifications. Its original outcome file contains 2,304 four-arm receiver rows. The input preparation script rebuilds the four execution payloads from pinned public TSV files; the matching script recreates the assignment with the pinned embedding snapshot. With the five licensed checkpoints available, the reproduction entry point executes sender extraction, alignment fitting, receiver interventions, and scoring. The package README gives these commands and their required inputs.

The \texttt{moral\_stories\_statistics/} component reconstructs Moral Stories statistics from released pair rows and donor assignments. It includes aggregate and secondary analyses, donor-overlap sensitivity code, and figure data. Its scope is released-row statistical reconstruction, not end-to-end model inference. The additional analyses in Appendix~\ref{app:reviewer-comparisons} retain separate configuration records and inferential families so that they can be distinguished from the original primary analyses.

\section{Responsible Research, Compute, and AI Assistance}
\label{app:responsible}

The five checkpoints contain approximately 3B, 7B, 8B, 12B, and 14B parameters. The original \mbox{XStoryCloze} campaign used one NVIDIA H200 at a time: confirmation required 17,664 forwards and 315 seconds; formation and instrumentation required 4,528, for 22,192 physical forwards overall.

Across the five added configurations, 11,520 receiver rows required 46,080 arm forwards. Baseline checks used 9,216 forwards and additional extraction used 1,920, for 57,216 overall. Higher-rank fitting reused the original formation activations, and rank-2 baseline margins reproduce the original rows exactly. This campaign has separate environment and run records. Historical Moral Stories runs lack a unified cumulative GPU-hour total.

No new human-participant study was conducted. The campaign coordinator performed the outcome-blind single-rater donor audit; it evaluated relatedness of existing story pairs. Generative AI use is disclosed in the Acknowledgments.

\end{document}